# Quality Estimation of English-Hindi Outputs using Naïve Bayes Classifier

Rashmi Gupta[#1], Nisheeth Joshi[#2], Iti Mathur[#3]

[#]Apaji Institute, Banasthali University, Rajasthan, India

[1]rsh.gupta06@gmail.com,
[2]nisheeth.joshi@rediffmail.com,
[3]mathur_iti@rediffmail.com,

*Abstract*— **In this paper we present an approach for estimating the quality of machine translation system. There are various methods for estimating the quality of output sentences, but in this paper we focus on Naïve Bayes classifier to build model using features which are extracted from the input sentences. These features are used for finding the likelihood of each of the sentences of the training data which are then further used for determining the scores of the test data. On the basis of these scores we determine the class labels of the test data.**

## I. INTRODUCTION

In this paper we have studied the confidence estimation for machine translation in which we find the quality scores of sentences. The main goal of confidence estimation is to judge the behaviour of the system output for a given input without any information about the expected output. Confidence Estimation for machine translation can be seen as a binary classification problem to distinguish between 'good' and 'bad' translations. The remaining section of the paper is organised as: Section 2 gives an overview of previous work on Quality Estimation. Section 3 describes the methodology of the system. Section 4 gives the Experimental settings. Section 5 gives tells us about the analysis and results.Section6 finally gives you the conclusion.

## II. RELATED WORK

Blatz.et al.[1] presents a number of experiments with CE at the sentence level based on annotations using automatic MT evaluation metrics. Repressors and classifiers are trained on features extracted for translations labeled according to National Institute of Standard and Technology (NIST) and Word Error Rate (WER). For classification, NIST scores are chosen to be threshold to label the 5th or 30th percentile of the examples as "good". For regression, the estimated scores are mapped into two classes using the same thresholds. The results did not show to be helpful in a range of evaluation tasks. Quirk [2] uses classifiers and a pre-defined threshold for "bad" and "good" translations considering a small set of 350 translations manually labeled for quality. Specia et al.[3] use a number of "black box" (MT system-independent) and "glass-box" (MT system-dependent) features to train a Partial Least Squares (PLS) regression algorithm to estimate both NIST and human scores. Gamon et al.[4] trained an SVM classifier using a number of linguistic features which were extracted from machine and human translations to differentiate between human and machine translations. Pado et.al.[5] used a regression algorithm along with features which contained textual entailment between the translation and the reference sentences. Soricut and Echihabi [6] focus on document level CE. The goal is to rank the documents according to their estimated quality and, given a threshold defined by the end-user, select the top n documents. Specia.et.al.[7] used 74 features to train a support vector machine classifier.

## III. METHODOLOGY

### A. Naive Bayes

Naïve Bayes is one of the most effective and efficient classification algorithms. In machine learning problems, a learner attempts to construct a classifier from a given set of training examples with class labels. Assume that $F_1$, $F_2$, $F_3$.., $F_n$ are n attributes. An example E is represented by a vector $(f_1, f_2, ....f_K)$, where $f_i$ is the value of $F_i$. Let C represent the class variable which takes values excellent, good, average and poor. We use QE to represent the value that C

takes. A naïve Bayesian classifier is defined as follows-

$$P(QE|F_1,F_2,F_3,F_4,\ldots F_N) = \frac{P(F1,F2,F3,.FN.|QE)*P(QE)}{P(F1,F2,F3,F4,\ldots FN)} \quad (1)$$

$$QE = \mathrm{argmax} P(F_1|QE)*P(F_2|QE)*P(F_3|QE).*P(F_N|QE)*P(QE) \quad (2)$$

Where,
P(QE|F) is the posterior probability of class (target) given predictor (attribute).
P(QE) is the prior probability of class.
P(F|QE) is the likelihood which is the probability of predictor given class.
P(F) is the prior probability of predictor.
The value of $(f_i/QE)$ can be estimated from the training example which can be easily implemented by Naïve Bayes classifier.

### B. Working of the system

Naïve Bayes is a well known algorithm for classification problems. The list of the sets of attribute values and its corresponding category are given to the classifier and these constitute the training set. From the training data an independent probability is established. The probability gives the likelihood of each target class, given the occurrence of each value category from each input variable. When a new example is presented a value for the target function can be predicted based on the training instances.

1 Training step- Using the training samples the methods estimate the parameter of a probability distribution, assuming features are conditionally independent given the class.

2 Testing step- for any unseen test sample, the method computes the posterior probability of that sample belongs to each class.

*Algorithmic steps*
- Input sentences
- Extract features from these input sentences.
- These features and their corresponding labels are provided to the learning algorithm.
- The model is trained using this learning algorithm (naive bayes classifier).
- Then raw data is taken and features are extracted from this raw data.
- This raw/ test data is provided to the trained model.
- We get the predicted label of the test data.

### IV. EXPERIMENTAL SETTINGS

#### A. Data Collection

For development of the training system, we used a 3,300 sentence corpus that was built during ACL 2005 workshop on building and using parallel text : Data Driven machine translation and beyond, as the training corpus. The statistics of the corpus is shown in table I

TABLE I
STATISTICS OF TRAINING CORPUS

| Corpus | English Hindi parallel corpus |
|---|---|
| Sentences | 3,300 |
| Words | 55,014 |
| Unique words | 8,956 |

#### B. Features Used

In this paper we have focused on using supervised machine learning in evaluation of MT engine outputs which does not use human reference translations. We have trained a Naïve Bayes classifier. In order to perform the task of confidence estimation across different machine translation systems we define a machine learning system which uses independent features which are extracted from the input (source) sentences and their corresponding translation (target) sentences. The set of 16 features are used in this paper are defined below-

- Number of token in the source sentence.
- Number of token in the target sentence.
- Average source token length.
- Language model probability of source sentence.
- Language model probability of target sentences.
- Average number of occurrence of target words within the target sentence.
- Average number of translation per source word in the sentence.
- Percentage of low frequency unigram in the source language.

- Percentage of high frequency unigram in the source language.
- Percentage of low frequency bigram in the source language.
- Percentage of high frequency bigram in the source language.
- Percentage of high frequency trigram in the source language.
- Percentage of low frequency trigram in the source language.
- Percentage of unigram in the source sentence seen in the corpus.
- Number of punctuation marks in the source sentence.
- Number of punctuation marks in the target sentence.

The outputs from training corpus are registered against all the three machine translation engines. The judging criteria was same as used by Joshi et. al. [8]. All the sentences are judged on ten parameters using a scale between 0-4 scores by human translators. The ten parameters used in the evaluations are as follows-

- Translation of Gender and Number of the Noun(s).
- Identification of the Proper Noun(s).
- Use of Adjectives and Adverbs corresponding to the Nouns and Verbs.
- Selection of proper words/synonyms (Lexical Choice).
- Sequence of phrases and clauses in the translation.
- Use of Punctuation Marks in the translation.
- Translation of tense in the sentence.
- Translation of Voice in the sentence.
- Maintaining the semantics of the source sentence in the translation.
- Fluency of translated text and translator's proficiency.

**Interpretation of human scale5**

- 1 = ideal
- 2 = perfect
- 3 = Acceptable
- 4 = Partially Acceptable
- 5 = Not Acceptable

Once the human evaluations of these outputs are done, we used these results along with the 16 features that were extracted from the English source sentences and Hindi MT outputs. We tested the classifiers using another corpus of 1300 sentences. The statistics of the test corpus is shown in table II.

TABLE II
STATISTICS OF TEST CORPUS

| corpus | English corpus |
|---|---|
| sentence | 1300 |
| words | 26,724 |
| Unique words | 3,515 |

The 1300 sentences were divided into 13 documents of 100 sentences each. We registered the outputs of the test corpus on all three machine translation engines and perform human evaluation on them.

V. ANALYSIS AND RESULTS

For the evaluation of the system we converted the human evaluation of the system into grades. These grades are given in the table III.

TABLE III
GRADE ALLOCATED TO HUMAN EVALUATIONS

| SNO | Score range | Grade |
|---|---|---|
| 1 | 0-0.250 | Poor |
| 2 | 0.251-0.50 | Average |
| 3 | 0.51-0.75 | Good |
| 4 | 0.751-1.0 | Excellent |

The results of the classifier are computed on the basis of these grades, which gave us the same class.

A. *Comparison of human evaluation and Naïve Bayes:*

The results produced by the naïve bayes classifier and the human evaluation are shown in table 4. This gives the number of times machine translation engine scored in each of the four categories. Table 5, shows the results of human evaluation for all the three MT engines. These four grades can also be converted into a numeric score to provide ranks to the MT outputs. Table 6 shows the comparison of results of human grades with the grades given by the classifier. If a human evaluator gave a good score to a sentence and the classifier also gave good to the same sentence then will counted it. More the human evaluator and the classifier can produce almost similar results to most of the judgments.

TABLE IV
NAIVE BAYES CLASSIFIER RESULTS

| SNO | Bing | Google | Babylon |
|---|---|---|---|
| Excellent | 24 | 23 | 12 |
| Good | 228 | 221 | 200 |
| Average | 1019 | 1008 | 1025 |
| Poor | 29 | 48 | 65 |

TABLE V
HUMAN EVALUATION RESULTS

| SNO | Bing | Google | Babylon |
|---|---|---|---|
| **Excellent** | 96 | 92 | 7 |
| **Good** | 231 | 194 | 234 |
| **Average** | 956 | 1002 | 1006 |
| **Poor** | 17 | 12 | 53 |

TABLE VI

COMPARISON OF NAIVE BAYES CLASSIFIER AND HUMAN EVALUATORS RESULTS

| SNO | Mt Engine | Same result | percentage |
|---|---|---|---|
| 1 | Bing | 771 | 59.30 |
| 2 | Google | 756 | 58.15 |
| 3 | Babylon | 711 | 54.69 |

VI. CONCLUSION AND FUTURE WORK

In this paper we have extracted 16 features from the input sentences and their translation and a quality score is obtained based on Bayesian inference produced from training data. In this paper we have shown that the Naïve Bayes classifier can predict the same level of outputs as that of human evaluator. Moreover in future we will improve the quality of the system by adding some more features and will study and evaluate the system in which we compare the rankings of the system given by the humans. The system which gives the same rank as given by the human will consider as the good system.

.